\documentclass[10pt,twocolumn,letterpaper]{article}

\usepackage[pagenumbers]{cvpr} % To force page numbers, e.g. for an arXiv version

\definecolor{cvprblue}{rgb}{0.21,0.49,0.74}
\usepackage[pagebackref,breaklinks,colorlinks,allcolors=cvprblue]{hyperref}

\def\paperID{9995} % *** Enter the Paper ID here
\def\confName{CVPR}
\def\confYear{2025}

\title{Probabilistic Prior Driven Attention Mechanism Based on Diﬀusion Model for Imaging Through Atmospheric Turbulence}

\author{Guodong Sun$^1$, Qixiang Ma$^2$, Liqiang Zhang$^1$, Hongwei Wang$^{1,}$\thanks{Corresponding author}, Zixuan Gao$^1$, Haotian Zhang$^1$\\
$^1$ Northwestern Polytechnical University\ \ \ $^2$ Beihang University\\
{\tt\small \{gavinsun0921, zhangliqiang2023, gaozixuan0909, 2022265247\}@mail.nwpu.edu.cn}\\
{\tt\small sycamore\_ma@buaa.edu.cn, wanghognwei@nwpu.edu.cn}
}

\begin{document}
\maketitle
\begin{abstract}
Atmospheric turbulence introduces severe spatial and geometric distortions, challenging traditional image restoration methods. We propose the Probabilistic Prior Turbulence Removal Network (PPTRN), which combines probabilistic diffusion-based prior modeling with Transformer-driven feature extraction to address this issue. PPTRN employs a two-stage approach: first, a latent encoder and Transformer are jointly trained on clear images to establish robust feature representations. Then, a Denoising Diffusion Probabilistic Model (DDPM) models prior distributions over latent vectors, guiding the Transformer in capturing diverse feature variations essential for restoration. A key innovation in PPTRN is the Probabilistic Prior Driven Cross Attention mechanism, which integrates the DDPM-generated prior with feature embeddings to reduce artifacts and enhance spatial coherence. Extensive experiments validate that PPTRN significantly improves restoration quality on turbulence-degraded images, setting a new benchmark in clarity and structural fidelity.
\end{abstract}    
\section{Introduction}
\label{sec:intro}

Imaging through atmospheric turbulence presents significant challenges due to severe and unpredictable distortions, including spatial and geometric aberrations that degrade image quality \cite{anantrasirichai2013atmospheric, mao2020image, lau2021atfacegan, zhu2012removing, yasarla2021learning, hirsch2010efficient, nair2021confidence, li2021unsupervised}. Common in applications like surveillance, astronomy, and remote sensing, these distortions obscure fine details, complicating reliable image restoration for high-fidelity analysis.

Traditional methods, including computational techniques and convolutional neural networks (CNNs), have been widely applied to mitigate turbulence-induced degradation \cite{park2020multi, zamir2021multi, dong2015image, ledig2017photo}. However, classical methods rely on simplified models unsuitable for dynamic conditions, while CNNs struggle to capture long-range dependencies essential for handling turbulence. As a result, these approaches often produce oversmoothed outputs lacking detail and structural consistency.

Restoring turbulence-degraded images is challenging due to high uncertainty and multi-modal distortions. Atmospheric turbulence requires probabilistic modeling to account for multiple plausible reconstructions and avoid oversmoothing \cite{rs16162972, jaiswal2023physics, yasarla2021learning, nair2021confidence}. Additionally, maintaining spatial coherence alongside fine details is critical, as traditional approaches often fail to balance these aspects, resulting in outputs that lack structural integrity or critical details \cite{mao2020image, anantrasirichai2013atmospheric, mao2022single}.

To address these challenges, we propose the Probabilistic Prior Turbulence Removal Network (PPTRN), a model that combines Denoising Diffusion Probabilistic Model (DDPM)-based probabilistic prior modeling with Transformer-driven feature extraction. The core of PPTRN is the Probabilistic Prior Driven Cross Attention mechanism, which fuses a DDPM-generated latent prior with feature embeddings to enhance detail preservation and spatial coherence. PPTRN employs a two-stage training strategy: initially, a latent encoder and Transformer are jointly trained on clear images; subsequently, the encoder’s weights are frozen, and DDPM models the prior distribution over latent vectors, guiding the Transformer in robust image restoration.

Our contributions include:
\begin{enumerate}
	\item A novel framework (PPTRN) that integrates probabilistic prior modeling with Transformer-based feature extraction for turbulence-distorted images.
	\item A Probabilistic Prior Driven Cross Attention mechanism that improves detail preservation and spatial coherence.
	\item A two-stage training strategy that balances structural consistency and detail preservation, capturing the multi-modal features of turbulence-affected images.
\end{enumerate}

These innovations highlight PPTRN’s effectiveness in restoring turbulence-degraded images, advancing uncertainty handling in visual restoration.

\begin{figure*}
  \centering
  \includegraphics[width=\linewidth]{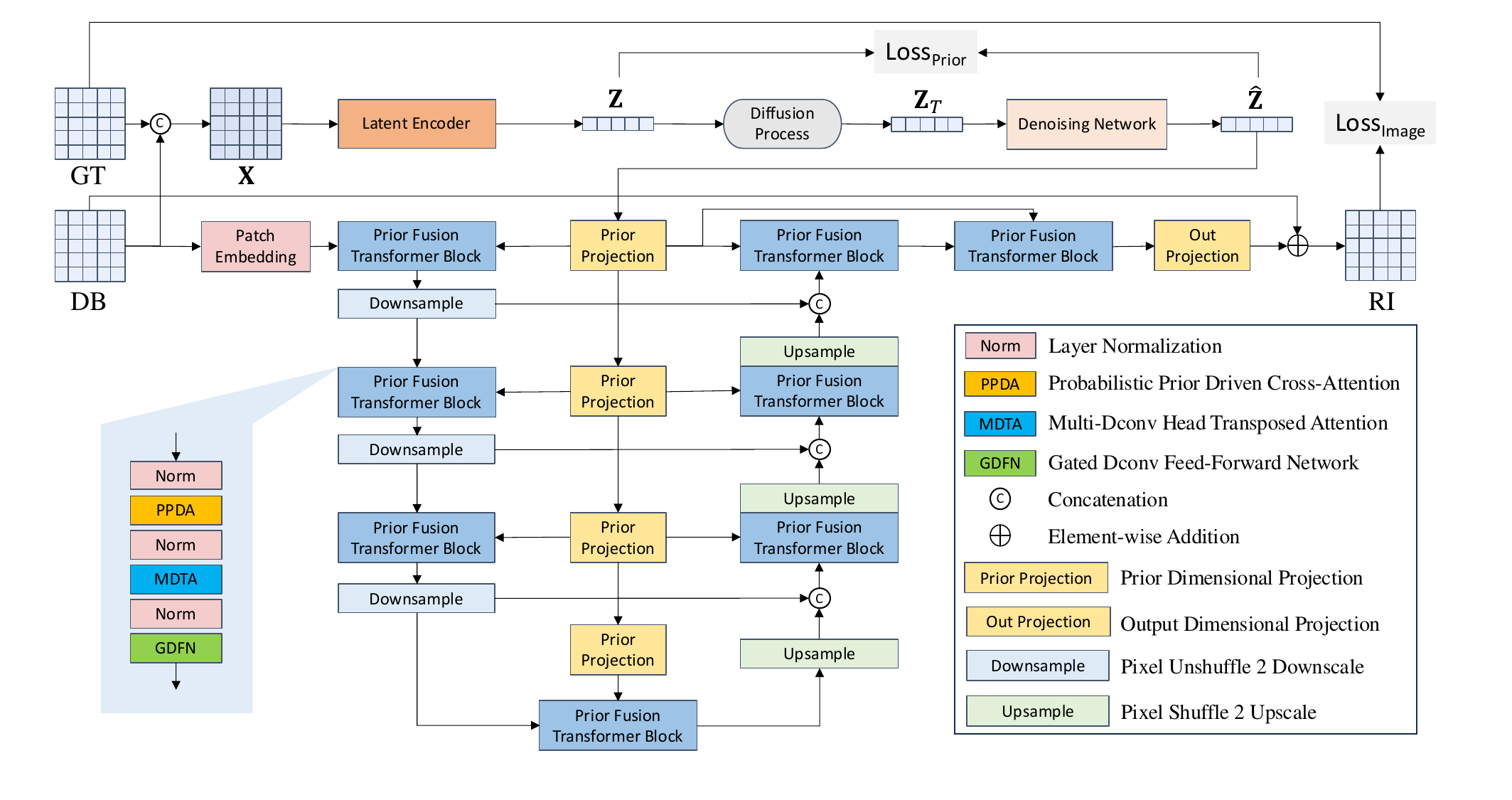}
  \caption{The network structure of the proposed Probabilistic Prior Turbulence Removal Network (PPTRN).}
  \label{fig:arch}
\end{figure*}

\section{Related Works}
\label{sec:related}

\subsection{Imaging Through Atmospheric Turbulence}

Restoring images affected by atmospheric turbulence has long been a challenge, particularly in remote sensing, surveillance, and astronomy applications \cite{rodriguez2022atmospheric, anantrasirichai2023atmospheric, mao2021accelerating, zhang2018removing}. Traditional methods, such as statistical modeling and optical flow, attempt to mitigate distortions by estimating spatial or temporal relationships between frames \cite{caliskan2014atmospheric, nieuwenhuizen2019dynamic, milanfar2012tour, zhang2017beyond}, but these approaches are limited by assumptions that don’t hold in dynamic turbulence conditions. Recently, deep learning methods, including CNNs and Transformers, have been explored for turbulence restoration \cite{mao2022single, tdrn, anantrasirichai2023atmospheric}. While CNNs are effective for local features, they struggle with long-range dependencies critical for turbulence. Transformers, although capable of capturing these dependencies, face challenges in handling the high uncertainty and multi-modal nature of turbulence-distorted images, underscoring the need for a hybrid approach.

\subsection{Diffusion Model for Image Restoration}

Diffusion models, inspired by non-equilibrium thermodynamics \cite{pmlr-v37-sohl-dickstein15}, have emerged as powerful generative models for representing complex, multi-modal data distributions. Denoising Diffusion Probabilistic Models (DDPMs) \cite{ho2020denoising} use iterative denoising to generate high-fidelity images, making them suitable for tasks involving significant uncertainty and multiple plausible outcomes. Recent studies have demonstrated diffusion models’ potential in image restoration, especially under high noise and complex distortions \cite{kawar2022denoising, xia2023diffir, saharia2022palette, saharia2022image, whang2022deblurring, song2019generative}. However, their application to atmospheric turbulence restoration remains limited. Our work leverages a DDPM-generated probabilistic prior to guide a Transformer-based architecture, enhancing the model’s ability to manage the high variability and structural inconsistencies inherent in turbulence, resulting in improved fidelity and detail preservation.

\section{Method}
\label{sec:method}

\begin{figure*}
  \centering
  \includegraphics[width=\linewidth]{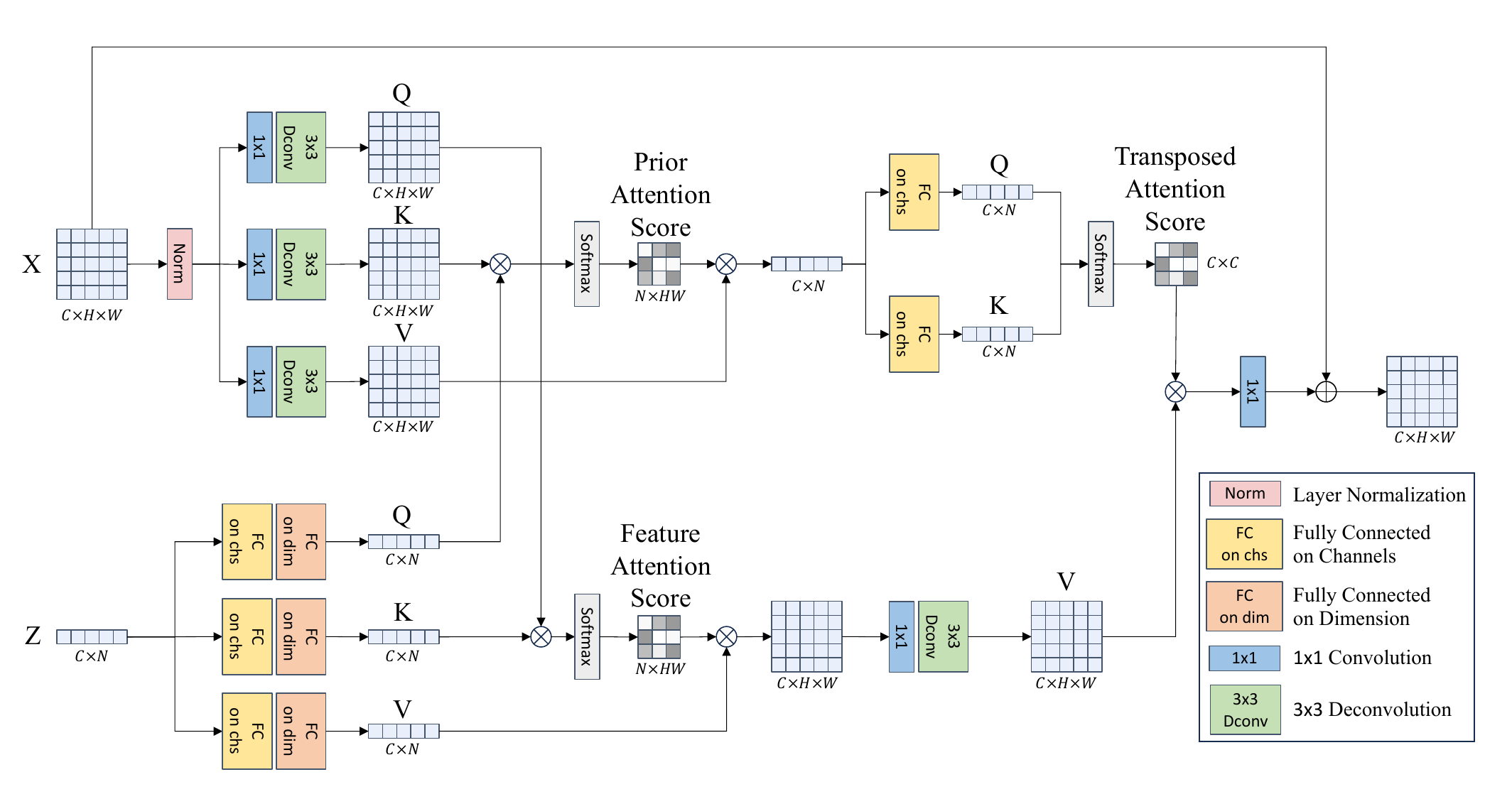}
  \caption{The structure of the proposed Probabilistic Prior Driven Cross-Attention (PPDA) mechanism.}
  \label{fig:cross_attention}
\end{figure*}

\subsection{Overview}

The Probabilistic Prior Turbulence Removal Network (PPTRN) is a novel image restoration model that mitigates atmospheric turbulence distortions by combining a Denoising Diffusion Probabilistic Model (DDPM) with Transformer-based feature extraction (see \cref{fig:arch}). In this framework, a probabilistic prior-driven cross-attention mechanism is employed, where DDPM generates a prior latent vector to capture clear image features, which the Transformer (Restormer) then utilizes to enhance spatial coherence and detail recovery. PPTRN is trained in two stages: first, the latent encoder and Transformer are jointly trained on clear images to establish stable feature representations; then, the encoder’s weights are frozen, and DDPM is introduced to model prior distributions that guide the Transformer in recovering details under complex turbulence conditions. This approach allows PPTRN to achieve robust restoration with improved structural fidelity and detail preservation.

\subsection{Model Components}

In this section, we describe the primary components of PPTRN: the latent encoder for initial feature representation, the Denoising Diffusion Probabilistic Model (DDPM) for prior modeling, and the Transformer for feature extraction. These modules work in tandem to provide a robust mechanism for restoring images degraded by atmospheric turbulence.

\textbf{Latent Encoder:} The Latent Encoder is a lightweight encoder designed to generate a concise representation of clear image features, which serves as the foundation for prior modeling. Given an input image $X$, the encoder maps it to a latent representation $Z$:
\begin{equation}
Z = f_{\text{encoder}}(X)
\end{equation}

This encoded representation $ Z $ captures essential features while preserving key details, and it is used as a fixed prior for DDPM in the second training stage. This latent encoding enables the Transformer to use probabilistic guidance during the image restoration process.

\textbf{Diffusion for Prior Modeling:} The Denoising Diffusion Probabilistic Model (DDPM) models a probabilistic prior distribution over the latent space, capturing the multi-modal characteristics of clear images. In the forward diffusion process, noise is gradually added to the latent representation through a Markov chain, defined as:
\begin{equation}
q(Z_t | Z_{t-1}) = \mathcal{N}(Z_t; \sqrt{1 - \beta_t} Z_{t-1}, \beta_t I)
\end{equation}
where $ \beta_t $ controls the noise level at each step. During the reverse diffusion process, DDPM progressively removes noise, yielding a prior latent vector $ \hat{Z} $ which guides the Transformer. This process is expressed as:
\begin{equation}
\hat{Z}{t-1} = \frac{1}{\sqrt{\alpha_t}} \left( Z_t - \frac{\beta_t}{\sqrt{1 - \alpha_t}} \epsilon\theta(Z_t, t) \right) + \sigma_t \epsilon
\end{equation}
where $ \alpha_t $ and $ \beta_t $ are diffusion process parameters, and $ \epsilon_\theta $ denotes the learned noise estimator. The prior vector $ \hat{Z} $ is thus enriched with uncertainty-aware features, providing critical guidance for the Transformer in restoring turbulence-degraded images.

\textbf{Transformer for Feature Extraction:} The Restormer Transformer is responsible for extracting detailed feature representations from turbulence-degraded images, using the DDPM-generated prior latent vector $ \hat{Z} $ within a cross-attention module. This cross-attention mechanism, explained in detail in \cref{sec:ppda}, enables the Transformer to selectively focus on relevant features by leveraging both local detail preservation and global coherence from the probabilistic prior. By integrating this prior information, the Transformer gains an enhanced contextual understanding, which helps mitigate spatial distortions and reconstruct fine-grained details effectively.

\subsection{Probabilistic Prior Driven Cross-Attention}
\label{sec:ppda}

The Probabilistic Prior Driven Cross Attention (PPDA) mechanism is a core innovation in PPTRN, designed to fuse the probabilistic prior latent vector from DDPM with the feature embeddings extracted by the Transformer (Restormer). By integrating prior information, PPDA enhances the Transformer’s ability to focus on spatial regions that require improved coherence and fine detail, facilitating effective restoration of turbulence-degraded images.

The structure of this cross-attention module is illustrated in \cref{fig:cross_attention}, showing the flow of information between the prior latent vector $ \mathbf{Z} $ and the feature embeddings $ \mathbf{X} $ extracted from the degraded input.

\textbf{Prior and Feature Input:} In the PPDA module, $ \mathbf{X} $ represents the feature embeddings derived from the degraded image, while $ \mathbf{Z} $ denotes the prior latent vector generated by DDPM. These inputs serve as guiding references to help the model reconstruct clear image structures. Before attention computation, both $ \mathbf{X} $ and $ \mathbf{Z} $ undergo initial transformations to align their feature distributions.

\textbf{Query, Key, and Value Generation:} To generate the Query ($ \mathbf{Q} $), Key ($ \mathbf{K} $), and Value ($ \mathbf{V} $) matrices, $ \mathbf{X} $ and $ \mathbf{Z} $ undergo distinct transformations:
\begin{itemize}
	\item For \textbf{Prior Attention} (using $ \mathbf{Z} $), the $ \mathbf{Q} $, $ \mathbf{K} $, $ \mathbf{V} $ representations are generated through fully connected layers:
\end{itemize}
\begin{equation}
	\mathbf{Q}_{\mathbf{Z}} = W_Q^{\mathbf{Z}} \mathbf{Z}, \quad \mathbf{K}_{\mathbf{Z}} = W_K^{\mathbf{Z}} \mathbf{Z}, \quad \mathbf{V}_{\mathbf{Z}} = W_V^{\mathbf{Z}} \mathbf{Z}
\end{equation}

\begin{itemize}
	\item For \textbf{Feature Attention} (using $ \mathbf{X} $), the transformations are applied using a sequence of 1x1 convolutions and 3x3 deconvolutions to produce:
\end{itemize}
\begin{equation}
\mathbf{Q}_{\mathbf{X}} = W_Q^{\mathbf{X}} \ast \mathbf{X}, \quad \mathbf{K}_{\mathbf{X}} = W_K^{\mathbf{X}} \ast \mathbf{X}, \quad \mathbf{V}_{\mathbf{X}} = W_V^{\mathbf{X}} \ast \mathbf{X}
\end{equation}
where $ W_Q^{\mathbf{Z}}, W_K^{\mathbf{Z}}, W_V^{\mathbf{Z}} $ are fully connected layers for $ \mathbf{Z} $, and $ W_Q^{\mathbf{X}}, W_K^{\mathbf{X}}, W_V^{\mathbf{X}} $ are the convolutional weights for $ \mathbf{X} $. This ensures compatibility between $ \mathbf{X} $ and $ \mathbf{Z} $ for effective interaction in the attention mechanism.

\textbf{Attention Score Computation:} PPDA computes two separate attention scores to capture both prior-guided and self-attention effects:
\begin{itemize}
	\item \textbf{Prior Attention Score}: This score leverages the prior knowledge from $ \mathbf{Z} $, calculated as:
\end{itemize}
\begin{equation}
A_{\text{prior}} = \text{softmax}\left(\frac{\mathbf{Q}_{\mathbf{Z}} \cdot \mathbf{K}_{\mathbf{X}}^\top}{\sqrt{d_k}}\right)
\end{equation}
\begin{itemize}
	\item \textbf{Feature Attention Score}: This score focuses on self-attention within $ \mathbf{X} $, computed as:
\end{itemize}
\begin{equation}
A_{\text{feature}} = \text{softmax}\left(\frac{\mathbf{Q}_{\mathbf{X}} \cdot \mathbf{K}_{\mathbf{Z}}^\top}{\sqrt{d_k}}\right)
\end{equation}

\textbf{Fusion and Recalculation of Query, Key, and Value:} After calculating the Prior Attention Score $ A_{\text{prior}} $ and Feature Attention Score $ A_{\text{feature}} $, these scores are used to fuse the initial Value representations $ \mathbf{V}_{\mathbf{Z}} $ and $ \mathbf{V}_{\mathbf{X}} $. A new set of Query, Key, and Value matrices for further refinement:
\begin{equation}
\begin{split}
	\mathbf{Q}^{\prime} &= W_Q^{\prime} A_{\text{prior}} \mathbf{V}_\mathbf{X} \\
	\mathbf{K}^{\prime} &= W_K^{\prime} A_{\text{prior}} \mathbf{V}_\mathbf{X} \\
	\mathbf{V}^{\prime} &= W_V^\prime A_{\text{feature}} \mathbf{V}_\mathbf{Z}
\end{split}
\end{equation}
where $ W_Q{\prime} $, $ W_K{\prime} $, and $ W_V{\prime} $ are learned weights for the recalculated Query, Key, and Value projections.

\textbf{Transposed Attention Score Computation:} The recalculated Query $ \mathbf{Q}^{\prime} $ and Key $ \mathbf{K}^{\prime} $ matrices are then used to compute the Transposed Attention Score $ A_{\text{transposed}} $, which further refines the alignment of fused representations. This score is calculated as follows:
\begin{equation}
	A_{\text{transposed}} = \text{softmax}\left(\frac{\mathbf{Q}’ \cdot \mathbf{K}’^\top}{\sqrt{d_c}}\right)
\end{equation}
where $ d_c $ represents the dimensionality of the channel space. This transposed attention mechanism allows for an additional layer of refinement, enhancing the model’s ability to capture intricate dependencies and spatial coherence in the image representation.

\textbf{Final Output Projection:} The Transposed Attention Score $ A_{\text{transposed}} $ is then applied to the recalculated Value $ \mathbf{V}{\prime} $ to produce the final refined output $ H_{\text{output}} $, which incorporates both prior-guided and feature-based attention for improved detail preservation:
\begin{equation}
	H_{\text{output}} = A_{\text{transposed}} \cdot \mathbf{V}’
\end{equation}

To prepare this fused representation for the next stage in the image reconstruction pipeline, $ H_{\text{output}} $ is passed through a 1x1 convolution layer to project it back to the original spatial dimensions. This ensures compatibility with subsequent processing stages and enhances the model’s ability to mitigate turbulence-induced distortions effectively.

In summary, the PPDA module integrates probabilistic prior information with feature embeddings through a multi-layered attention mechanism, enabling the model to dynamically focus on critical regions and improve the fidelity and structural integrity of the restored image.

\subsection{Two-Stage Training Strategy}

The Two-Stage Training Strategy is essential for optimizing PPTRN’s performance by sequentially leveraging the latent encoder and diffusion-based prior modeling. This approach allows the model to first establish stable feature representations and then refine them through probabilistic modeling, enhancing its robustness in handling complex atmospheric distortions.

\textbf{Stage 1: Joint Training of Latent Encoder and Transformer} In the first stage, the latent encoder and Transformer are jointly trained on a dataset of clear images. The latent encoder generates a latent vector that captures essential structural information, forming a foundational representation that guides feature extraction. During this phase, the Transformer learns to process the encoder’s latent vector alongside degraded image features, optimizing feature extraction and cross-attention based on the characteristics of clean images. This joint training enables the model to build a strong initial understanding of image structures and key details.

\textbf{Stage 2: Prior Modeling with DDPM and Transformer Fine-Tuning} In the second stage, the weights of the latent encoder are frozen to retain the clear image characteristics learned in Stage 1. The Denoising Diffusion Probabilistic Model (DDPM) is then introduced to model a probabilistic prior over this fixed latent space, producing prior latent vectors that reflect the multi-modal nature of clear images. The DDPM-generated prior guides the Transformer in restoring turbulence-degraded images, helping it to adapt to diverse image attributes and reduce the risk of generating oversmoothed results.

This two-stage strategy balances stability and flexibility, allowing PPTRN to capture stable representations of clear images while benefiting from the diversity introduced by probabilistic modeling. Together, these stages enhance PPTRN’s ability to generate high-fidelity, detail-rich restorations under challenging conditions.

\section{Experiments}
\label{sec:experiments}

\subsection{Implementation}

\textbf{Traning Details} The Probabilistic Prior Turbulence Removal Network (PPTRN) was trained on the Atmospheric Turbulence Distorted Video Sequence Dataset (\textbf{ATDVSD}) \cite{jin2021neutralizing} using the AdamW optimizer with a learning rate of $1 \times 10^{-4}$ and a batch size of 16. Training was conducted for 400,000 steps with early stopping based on validation loss to prevent overfitting. A cosine variance schedule was applied to the DDPM component for smooth noise control across diffusion steps. The Transformer architecture in PPTRN, adapted from Restormer, was modified to support the probabilistic prior-driven cross-attention mechanism.

\noindent\textbf{Datasets}

\begin{itemize}
	\item \textbf{ATDVSD} \cite{jin2021neutralizing}: This dataset includes video sequences of scenes distorted by atmospheric turbulence, converted into single-frame images for frame-based restoration. It combines physical (environment-controlled) and algorithmic (computationally simulated) turbulence, offering diverse examples for training.
	\item \textbf{Heat Chamber Dataset} \cite{mao2022single}: Collected by introducing heat along the optical path to simulate atmospheric turbulence, this dataset contains realistic distortion patterns similar to outdoor turbulence. While not used in training, it served as an evaluation benchmark to test PPTRN’s generalization to physically simulated turbulence.
\end{itemize}

\noindent\textbf{Implementation Environment} Experiments were conducted on NVIDIA RTX 3090 GPUs using PyTorch (v2.4.1). Training and inference were parallelized with Hugging Face Accelerate \cite{accelerate} to speed up processing. Hyperparameter tuning and evaluations were automated to ensure reproducibility.

\subsection{Quantitative Evaluation}

To assess the effectiveness of the Probabilistic Prior Turbulence Removal Network (PPTRN) in restoring turbulence-degraded images, we conducted experiments on the Atmospheric Turbulence Distorted Video Sequence Dataset (ATDVSD) \cite{jin2021neutralizing} and the Heat Chamber Dataset \cite{mao2022single}. We evaluated performance using three metrics—PSNR (Peak Signal-to-Noise Ratio), SSIM (Structural Similarity Index) \cite{wang2004image}, and LPIPS (Learned Perceptual Image Patch Similarity) \cite{zhang2018unreasonable}—to measure restoration fidelity, structural consistency, and perceptual quality.

\noindent\textbf{ATDVSD} \cite{jin2021neutralizing} For ATDVSD, we tested PPTRN on both physically simulated and algorithmically simulated turbulence images, as shown in Table \ref{tab:atmospheric_results}. These two types of simulations represent different characteristics of atmospheric turbulence, providing a comprehensive test for PPTRN’s adaptability.

\begin{table}
    \centering
%    \resizebox{\linewidth}{!}{
    \begin{tabular}{@{}ccccc@{}}
        \toprule
        Method & Type & PSNR$\uparrow$ & SSIM$\uparrow$ & LPIPS$\downarrow$ \\ 
        \midrule
        TDRN \cite{tdrn} & Phys & 32.0029 & 0.8894 & 0.1907 \\ 
        MTRNN \cite{park2020multi} & Phys & 31.7379 & 0.8947 & 0.1885 \\ 
        MPRNet \cite{zamir2021multi} & Phys & 32.6535 & 0.9059 & 0.1819 \\ 
        Uformer \cite{wang2022uformer} & Phys & 35.8975 & 0.9297 & 0.1512 \\ 
        Restormer \cite{zamir2022restormer} & Phys & 36.5793 & 0.9519 & \underline{0.1319} \\ 
        Stripformer \cite{tsai2022stripformer} & Phys & 36.3699 & 0.9499 & 0.1356 \\ 
        TurbNet \cite{mao2022single} & Phys & \underline{36.5826} & \underline{0.9563} & 0.1332 \\ 
        PPTRN & Phys & \textbf{37.1898} & \textbf{0.9562} & \textbf{0.1261} \\ 
        \midrule
        TDRN \cite{tdrn} & Algo & 30.1149 & 0.9089 & 0.2023 \\ 
        MTRNN \cite{park2020multi} & Algo & 28.8190 & 0.8906 & 0.1942 \\ 
        MPRNet \cite{zamir2021multi} & Algo & 31.6241 & 0.9099 & 0.1811 \\ 
        Uformer \cite{wang2022uformer} & Algo & 34.7358 & 0.9384 & 0.1478 \\ 
        Restormer \cite{zamir2022restormer} & Algo & 36.4814 & \underline{0.9531} & \underline{0.1277} \\ 
        Stripformer \cite{tsai2022stripformer} & Algo & 35.6583 & 0.9459 & 0.1348 \\ 
        TurbNet \cite{mao2022single} & Algo & \underline{36.7162} & 0.9473 & 0.1281 \\ 
        PPTRN & Algo & \textbf{37.2929} & \textbf{0.9535} & \textbf{0.1273} \\ 
        \bottomrule
    \end{tabular}
%    }
    \caption{Quantitative results on the Atmospheric Turbulence Distorted Video Sequence Dataset. Type column abbreviations: Phys denotes methods tested on physically simulated data, and Algo denotes methods tested on algorithmically simulated data.}
    \label{tab:atmospheric_results}
\end{table}

As shown in Table \ref{tab:atmospheric_results}, PPTRN achieves superior PSNR and SSIM scores across both turbulence types, indicating better fidelity and structural consistency compared to baseline methods. Additionally, PPTRN’s lower LPIPS scores highlight its ability to produce perceptually more realistic restorations with fewer artifacts.

\noindent\textbf{Heat Chamber Dataset} \cite{mao2022single} To test the generalization capability of PPTRN, we evaluated it on the Heat Chamber Dataset, which simulates atmospheric turbulence by introducing heat along the optical path. This dataset was not used during training, providing an independent test for model robustness. Table \ref{tab:heat_chamber_results} shows that PPTRN outperforms baseline models, achieving higher PSNR and SSIM values and lower LPIPS scores, confirming its adaptability to diverse turbulence conditions.

\begin{table}
    \centering
    \begin{tabular}{@{}cccc@{}}
        \toprule
        Method & PSNR$\uparrow$ & SSIM$\uparrow$ & LPIPS$\downarrow$ \\ 
        \midrule
        TDRN \cite{tdrn} & 18.4267 & 0.6424 & 0.3713 \\ 
        MTRNN \cite{park2020multi} & 18.3734 & 0.6379 & 0.3981 \\ 
        MPRNet \cite{zamir2021multi} & 18.6871 & 0.6519 & 0.3774 \\ 
        Uformer \cite{wang2022uformer} & 19.0327 & 0.6638 & 0.3679 \\ 
        Restormer \cite{zamir2022restormer} & \underline{19.3237} & 0.6741 & \underline{0.3531} \\ 
        Stripformer \cite{tsai2022stripformer} & 19.1982 & 0.6648 & 0.3628 \\ 
        TurbNet \cite{mao2022single} & 19.3186 & \textbf{0.6812} & 0.3533 \\ 
        PPTRN & \textbf{19.4260} & \underline{0.6793} & \textbf{0.3440} \\ 
        \bottomrule
    \end{tabular}
    \caption{Quantitative results on the Heat Chamber Dataset.}
    \label{tab:heat_chamber_results}
\end{table}

Table \ref{tab:heat_chamber_results} demonstrates PPTRN’s ability to maintain high-quality restoration across varying types of turbulence, affirming its robustness and generalization potential.

\subsection{Qualitative Comparisons}

\begin{figure}
    \centering
    \resizebox{\linewidth}{!}{
        \begin{minipage}{\linewidth}
            \centering
            \begin{minipage}{0.31\linewidth}
                \centering
                \includegraphics[width=\linewidth]{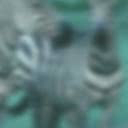}
                \vspace{-7mm}
                \caption*{Input}
            \end{minipage}
            \begin{minipage}{0.31\linewidth}
                \centering
                \includegraphics[width=\linewidth]{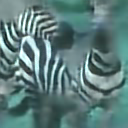}
                \vspace{-7mm}
                \caption*{MTRNN}
            \end{minipage}
            \begin{minipage}{0.31\linewidth}
                \centering
                \includegraphics[width=\linewidth]{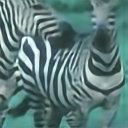}
                \vspace{-7mm}
                \caption*{MPRNet}
            \end{minipage}

            \begin{minipage}{0.31\linewidth}
                \centering
                \includegraphics[width=\linewidth]{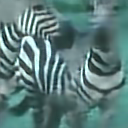}
                \vspace{-7mm}
                \caption*{Uformer}
            \end{minipage}
            \begin{minipage}{0.31\linewidth}
                \centering
                \includegraphics[width=\linewidth]{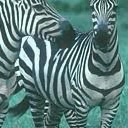}
                \vspace{-7mm}
                \caption*{Stripformer}
            \end{minipage}
            \begin{minipage}{0.31\linewidth}
                \centering
                \includegraphics[width=\linewidth]{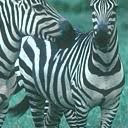}
                \vspace{-7mm}
                \caption*{Restormer}
            \end{minipage}

            \begin{minipage}{0.31\linewidth}
                \centering
                \includegraphics[width=\linewidth]{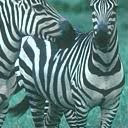}
                \vspace{-7mm}
                \caption*{TurbNet}
            \end{minipage}
            \begin{minipage}{0.31\linewidth}
                \centering
                \includegraphics[width=\linewidth]{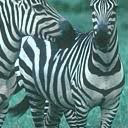}
                \vspace{-7mm}
                \caption*{PPTRN (Ours)}
            \end{minipage}
            \begin{minipage}{0.31\linewidth}
                \centering
                \includegraphics[width=\linewidth]{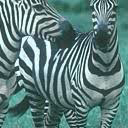}
                \vspace{-7mm}
                \caption*{Reference}
            \end{minipage}
        \end{minipage}
    }
    \caption{Visual comparison of image restoration results on a sample image with various models. From left to right: Input (degraded image), results from MTRNN, MPRNet, Uformer, Stripformer, Restormer, TurbNet, our proposed PPTRN, and the Reference image. PPTRN (Ours) shows superior restoration quality, closely matching the reference with minimal artifacts.}
    \label{fig:visualize_1}
\end{figure}

To further evaluate PPTRN’s effectiveness, we present qualitative comparisons with baseline models. \cref{fig:visualize_1} shows visual comparisons of image restoration results on a sample image across various models, including MTRNN, MPRNet, Uformer, Stripformer, Restormer, TurbNet, and our proposed PPTRN, along with the reference image. From left to right, the progression highlights PPTRN’s superior ability to restore fine details and maintain structural coherence, closely matching the reference with minimal artifacts. Unlike other methods, PPTRN produces clear, artifact-free images, enhancing perceptual fidelity even in regions with significant turbulence-induced distortions.

\begin{figure}
    \centering
    \resizebox{\linewidth}{!}{
        \begin{minipage}{\linewidth}
            \centering
            \begin{minipage}{0.31\linewidth}
                \centering
                \includegraphics[width=\linewidth]{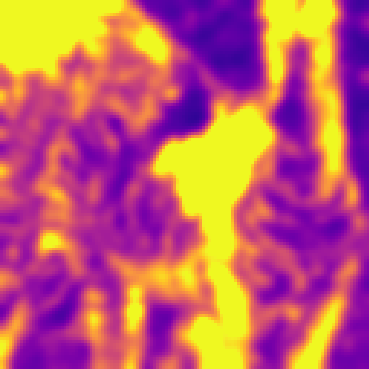}
                \vspace{-7mm}
                \caption*{MTRNN}
            \end{minipage}
            \begin{minipage}{0.31\linewidth}
                \centering
                \includegraphics[width=\linewidth]{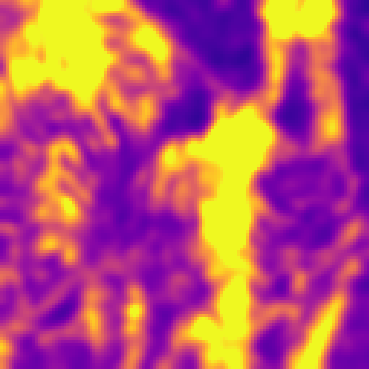}
                \vspace{-7mm}
                \caption*{MPRNet}
            \end{minipage}
            \begin{minipage}{0.31\linewidth}
                \centering
                \includegraphics[width=\linewidth]{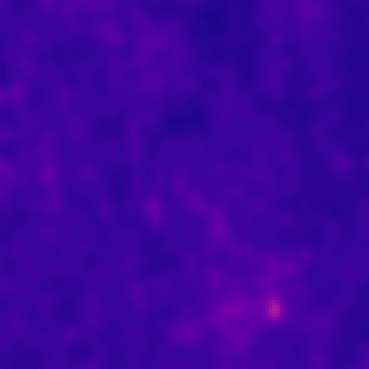}
                \vspace{-7mm}
                \caption*{Stripformer}
            \end{minipage}

            \begin{minipage}{0.31\linewidth}
                \centering
                \includegraphics[width=\linewidth]{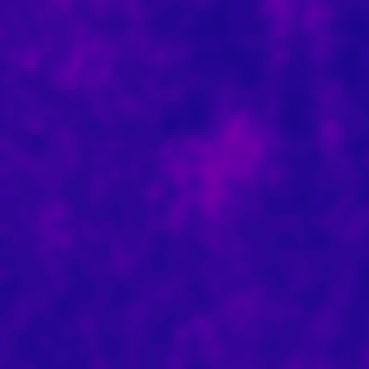}
                \vspace{-7mm}
                \caption*{Restormer}
            \end{minipage}
            \begin{minipage}{0.31\linewidth}
                \centering
                \includegraphics[width=\linewidth]{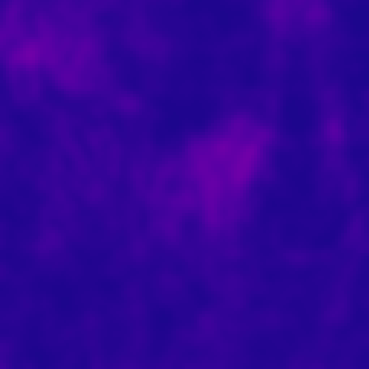}
                \vspace{-7mm}
                \caption*{TurbNet}
            \end{minipage}
            \begin{minipage}{0.31\linewidth}
                \centering
                \includegraphics[width=\linewidth]{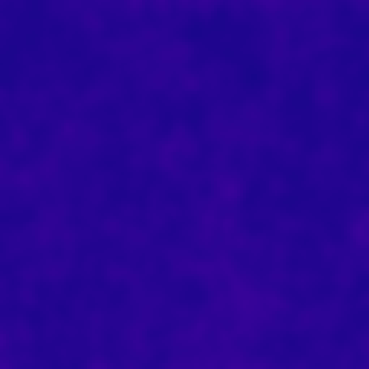}
                \vspace{-7mm}
                \caption*{PPTRN (Ours)}
            \end{minipage}
        \end{minipage}
    }
    \caption{Error maps of different models compared to the reference image. Lower error intensities indicate better restoration quality. PPTRN (Ours) demonstrates the lowest error, highlighting its effectiveness in reducing distortion and achieving accurate image reconstruction.}
    \label{fig:visualize_2}
\end{figure}

\cref{fig:visualize_2} provides error maps for different models, comparing each model’s restoration to the reference image. Lower error intensities in these maps represent better restoration quality. PPTRN achieves the lowest error levels, demonstrating its effectiveness in reducing distortion and accurately reconstructing image details. This highlights PPTRN’s robustness in preserving both local details and global structure, reducing artifacts more effectively than competing methods.

Together, \cref{fig:visualize_1} and \cref{fig:visualize_2} underscore PPTRN’s advantage in qualitative performance, showing its ability to generate high-quality restorations that align closely with the reference both visually and structurally.

\subsection{Ablation Studies}

To further analyze the contribution of each component in PPTRN, we conducted a series of ablation studies. These experiments evaluate the impact of the probabilistic prior-driven cross-attention mechanism and the two-stage training strategy on model performance.

\textbf{Effect of Probabilistic Prior Driven Cross Attention} In this experiment, we removed the probabilistic prior-driven cross-attention mechanism from PPTRN, using only the baseline Transformer architecture, comparable to Restormer. As shown in Table \ref{tab:cross_attention_ablation}, removing the cross-attention module led to a drop in PSNR and SSIM, along with an increase in LPIPS. This result highlights the importance of integrating the DDPM-generated prior with Transformer features, as it significantly enhances detail preservation and spatial coherence in turbulence-distorted images.

\begin{table}
    \centering
    \begin{tabular}{@{}cccc@{}}
        \toprule
        Model & PSNR$\uparrow$ & SSIM$\uparrow$ & LPIPS$\downarrow$ \\ 
        \midrule
        Without Cross Attention & 36.4814 & 0.9531 & 0.1277 \\ 
        PPTRN (ours) & \textbf{37.2929} & \textbf{0.9535} & \textbf{0.1273} \\ 
        \bottomrule
    \end{tabular}
    \caption{Ablation study results showing the impact of the Probabilistic Prior Driven Cross Attention on model performance. The inclusion of cross attention improves PSNR and SSIM, with a slight decrease in LPIPS, highlighting its effectiveness in enhancing image quality and structural consistency.}
    \label{tab:cross_attention_ablation}
\end{table}

\textbf{Effect of Joint Module Training} To assess the impact of joint module training within the two-stage training strategy, we conducted an experiment in which, during the second stage, only the DDPM was trained to model the prior information encoded by the latent encoder, without jointly training the Transformer. As seen in Table \ref{tab:two_stage_ablation}, the full two-stage strategy, where both the DDPM and Transformer are jointly optimized, results in superior performance. This demonstrates that the simultaneous training of both components allows the model to more effectively utilize prior information, leading to improved restoration quality.

\begin{table}
    \centering
    \begin{tabular}{@{}cccc@{}}
        \toprule
        Model & PSNR$\uparrow$ & SSIM$\uparrow$ & LPIPS$\downarrow$ \\ 
        \midrule
        PPTRN w/o joint & 36.3712 & 0.9417 & 0.1433 \\ 
        PPTRN & \textbf{37.2929} & \textbf{0.9535} & \textbf{0.1273} \\ 
        \bottomrule
    \end{tabular}
    \caption{Ablation study results on the influence of the Two-Stage Training Strategy. Joint training in the two-stage approach significantly improves PSNR and SSIM while reducing LPIPS, demonstrating its effectiveness in enhancing image quality and structural fidelity.}
    \label{tab:two_stage_ablation}
\end{table}

These ablation studies confirm the critical role of each component in PPTRN. The probabilistic prior-driven cross-attention mechanism (Table \ref{tab:cross_attention_ablation}) and the joint training strategy in the two-stage process (Table \ref{tab:two_stage_ablation}) collectively enhance PPTRN’s capability to restore high-quality images from turbulence-distorted inputs, validating the effectiveness of our architectural and training design choices.

\section{Conclusion}
\label{sec:conclusion}

We presented the Probabilistic Prior Turbulence Removal Network (PPTRN), a novel model that integrates diffusion-based prior modeling with a Transformer framework to address the multi-modal and complex distortions caused by atmospheric turbulence. By leveraging a two-stage training strategy and a Probabilistic Prior Driven Cross Attention mechanism, PPTRN effectively combines probabilistic prior information with feature embeddings, allowing for the restoration of fine details while preserving spatial coherence. Experimental results on the Atmospheric Turbulence Distorted Video Sequence Dataset and Heat Chamber Dataset demonstrate PPTRN’s superiority in image clarity and structural fidelity, setting a new benchmark in turbulence-degraded image restoration. Ablation studies further validate the contributions of each component, highlighting the effectiveness of integrating DDPM-based priors with a Transformer architecture. This work underscores the potential of probabilistic modeling in enhancing restoration performance in challenging atmospheric conditions and opens up new possibilities for applying such approaches to other complex image restoration tasks where uncertainty and structural integrity are critical.
{
    \small
    \bibliographystyle{ieeenat_fullname}
    \bibliography{main}
}

% WARNING: do not forget to delete the supplementary pages from your submission 
% \input{sec/X_suppl}

\end{document}